\documentclass[11pt,oneside]{article}

\usepackage[a4paper,
            bindingoffset=0.2in,
            left=1in,
            right=1in,
            top=1in,
            bottom=1in,
            footskip=.25in]{geometry}

\usepackage[english]{babel}
\usepackage[utf8]{inputenc}
\usepackage{amsmath,amsfonts}
\usepackage{booktabs}
\usepackage{natbib}
\usepackage{framed,multirow}
\usepackage[table]{xcolor}
\usepackage{xspace}
\usepackage{subfig}
\usepackage{authblk}
\usepackage{graphicx}
\usepackage{url}
\usepackage{soul}

\newcommand{\bm}[1]{\mathbf{#1}}
\makeatletter
\newcommand*\bigcdot{\mathpalette\bigcdot@{.5}}
\newcommand*\bigcdot@[2]{\mathbin{\vcenter{\hbox{\scalebox{#2}{$\m@th#1\bullet$}}}}}
\makeatother

\title{AI-Driven Early Mental Health Screening: Analyzing Selfies of Pregnant Women}

\author[1]{Gustavo A. Basílio}
\author[1]{Thiago B. Pereira}
\author[2]{Alessandro L. Koerich}
\author[3]{Hermano Tavares}
\author[1]{Ludmila Dias}
\author[4]{Maria das Graças da S. Teixeira}
\author[5]{Rafael T. Sousa}
\author[4]{Wilian H. Hisatugu}
\author[3]{Amanda S. Mota}
\author[6]{Anilton S. Garcia}
\author[7]{Marco Aurélio K. Galletta}
\author[1]{Thiago M. Paixão}
\affil[1]{{\small Federal Institute of Espírito Santo (IFES), Campus Serra, Serra, Brazil}}
\affil[2]{{\small École de Technologie Supérieure (ÉTS), Montreal, Canada}}
\affil[3]{{\small Department of Psychiatry, University of São Paulo Medical School
(FMUSP), São Paulo, Brazil}}
\affil[4]{{\small Department of Computing and Electronics, Federal University of Espírito Santo (UFES), Campus São Mateus, São Mateus, Brazil}}
\affil[5]{{\small Federal University of Mato Grosso (UFMT), Barra do Garças, Brazil}}
\affil[6]{{\small Federal University of Espírito Santo (UFES), Campus Goiabeiras, Vitória, Brazil}}
\affil[7]{{\small Department of Obstetrics and Gynecology, University of São Paulo Medical School
(FMUSP), São Paulo, Brazil}}

\date{}

\begin{document}
\maketitle

\begin{abstract}
Major Depressive Disorder and anxiety disorders affect millions globally, contributing significantly to the burden of mental health issues. Early screening is crucial for effective intervention, as timely identification of mental health issues can significantly improve treatment outcomes. Artificial intelligence (AI) can be valuable for improving the screening of mental disorders, enabling early intervention and better treatment outcomes. AI-driven screening can leverage the analysis of multiple data sources, including facial features in digital images. However, existing methods often rely on controlled environments or specialized equipment, limiting their broad applicability. This study explores the potential of AI models for ubiquitous depression-anxiety screening given face-centric selfies. The investigation focuses on high-risk pregnant patients, a population that is particularly vulnerable to mental health issues. To cope with limited training data resulting from our clinical setup, pre-trained models were utilized in two different approaches: fine-tuning convolutional neural networks (CNNs) originally designed for facial expression recognition and employing vision-language models (VLMs) for zero-shot analysis of facial expressions. Experimental results indicate that the proposed VLM-based method significantly outperforms CNNs, achieving an accuracy of 77.6\%. Although there is significant room for improvement, the results suggest that VLMs can be a promising approach for mental health screening.
\end{abstract}




\maketitle

\section{Introduction}

Major depressive disorder (depression) is a prevalent mental health disorder affecting over 280 million people globally and a leading cause of disability worldwide. It contributes significantly to the global disease burden \citep{WHO2023}. Similarly, anxiety disorders, marked by excessive worry and fear, impact millions of lives, often leading to debilitating effects on daily functioning and quality of life. Particularly, there is a recognized vulnerability to mood disorders during pregnancy, with possible worsening of previous emotional conditions and the emergence of new altered mental states that increase the risk of depression and anxiety \citep{biaggi2016identifying}. This vulnerability is certainly influenced by the significant hormonal changes of pregnancy. However, it is also important to recognize that this is a time of great physical, psycho-emotional, cultural, and social changes, generating psychic stress and increased anxiety.

Several studies have pointed to the association between depression and anxiety during pregnancy with unfavorable obstetric and neonatal outcomes. These disorders can increase the risk of obstetric complications such as cesarean delivery, preeclampsia, preterm birth, low birth weight, small for gestational age newborns, and newborns with low Apgar scores, indicating lower oxygenation at birth \citep{li2021mood,nasreen2019impact,dowse2020impact,kurki2000depression}. These findings underline the importance of making a correct and early diagnosis, allowing for appropriate psychiatric and psychological follow-up during pregnancy to improve both maternal and neonatal outcomes.

For early intervention and improved treatment outcomes, screening processes are essential for identifying individuals who may have conditions such as depression or anxiety disorders before they present symptoms or seek treatment \citep{thombs2023screening}. Various methods can be employed for screening, including self-reported questionnaires, clinical interviews with mental health professionals, and observing behaviors and physical symptoms. Additionally, studies in affective computing, a field that explores the interaction between human emotions and computational systems, have demonstrated the potential of artificial intelligence (AI) in screening mental disorders \citep{kumar2024measuring}. AI-driven technologies can enhance screening by analyzing data from multiple sources, such as text inputs, voice recordings, or facial expressions \citep{liu2024}, which is the very focus of this work.

In this work, we address the challenge of ubiquitous depression-anxiety screening from face-centric selfie images in a real-world clinical setting involving high-risk pregnant patients. This effort is part of a broader research project led by our group, aimed at developing mobile applications for mental health assessment. In our application scenario, the user takes a selfie with a smartphone front-facing camera. The application sends the image to a server, where an AI model analyzes it and provides a label indicating whether the patient is normal or has symptoms of depression-anxiety. The server returns the label to the application, providing feedback to the user.
To train and evaluate the models, we use image data (selfies) and responses to the Patient Health Questionnaire-4 (PHQ-4), a brief screening tool for anxiety and depression. The PHQ-4 responses are used to derive image labels -- normal or abnormal (depression-anxiety) --, which play the role of supervisory signals during training and ground truth for evaluation. To the best of our knowledge, this is the first study that investigates the use of face-centric selfies for screening anxiety and depression in high-risk pregnant patients in a clinical context.

In line with the state-of-the-art, deep models are employed in facial analysis. Pre-trained models are leveraged using two distinct approaches to address the challenge of limited training data (most participants contributed with only a single photo). In a more traditional approach, we employ a transfer learning strategy where convolutional neural networks (CNNs) originally trained for facial expression recognition are fine-tuned for depression-anxiety detection. In a more innovative approach, we propose using powerful vision-language models (VLMs) as a facial analyzer. While VLMs are not suitable for directly assessing depression-anxiety, as demonstrated in our experiments, they excel in zero-shot detailed descriptions of facial expressions that correspond to basic emotions (e.g., anger, happiness, and sadness). This approach detects depression-anxiety by classifying the generated text with simple neural networks instead of directly classifying the image. Experiments conducted under a rigid \emph{Leave One Subject Out} protocol revealed that our VLM-based approach outperformed the traditional CNNs, achieving an accuracy of 77.6\%, which represents a gain of approx. 10.0 percentage points (p.p.) compared to the CNNs, and an F1-score of 56.0\%, an improvement of approx. 11.0 p.p..

In summary, the main contributions of this work are:

\begin{itemize}
    \item A novel VLM-based approach for ubiquitous mental-health screening from selfies.
    \item A study on the use of AI for depression-anxiety screening in high-risk pregnant patients.
    \item Collection of a dataset\footnote{Anonymized descriptions produced by the VLMs and corresponding PHQ-4 responses are publicly available at \url{https://bit.ly/3E1EPKw}.} comprising selfies and PHQ-4 responses from high-risk pregnant patients.
    \item Comprehensive assessment of VLMs and CNNs for depression-anxiety screening with limited data.
\end{itemize}

The rest of this paper is organized as follows. The next section discusses the related works. Section \ref{sec:method} describes the two approaches for AI-driven screening methodology. Section \ref{sec:experimental-methodology} presents the experimental setup, while Section \ref{sec:results} discusses the results. Finally, Section \ref{sec:conclusion} concludes the paper and discusses future work opportunities.
\section{Related Work}

The use of facial features to identify non-basic emotions (depression, stress, engagement, shame, guilt, envy, among others) is becoming increasingly popular in various applications. \cite{nepal2024moodcapture} highlight several elements explored in this type of analysis, such as facial expression itself, gaze direction, as well as general image characteristics like luminosity and background settings. In this domain, machine learning (ML) plays an essential role, particularly with the use of deep learning (DL) models and related techniques, such as transfer learning and attention mechanisms. As highlighted by \cite{kumar2024measuring}, DL models have become the primary choice for detecting non-basic emotions since 2020, surpassing traditional machine learning and image processing algorithms. Regarding models, convolutional neural networks (CNNs) are the most popular choice for image-based analysis of non-basic emotions. For instance, \cite{gupta2023facial} used CNNs to quantify student engagement in online learning, while \cite{zhou2018visually} applied them for detecting depression based on facial images. Transfer learning plays a crucial role as pre-trained models for tasks like basic facial expression recognition can be fine-tuned for specific targets like detecting stress \citep{voleti2024stress}. Usually, pre-training leverages larger datasets, which is particularly useful when the target dataset is small, as is often the case in mental health applications. Attention mechanisms are also relevant in this context, as they can help models focus on specific regions of the face that are more informative for the target task \citep{viegas2018towards,belharbi2024guided}.

Despite the promising results, most studies on depression detection through facial expressions involve capturing images in controlled environments where individuals follow a predetermined script, which limits the realism of the facial expressions \citep{liu2022measuring,kong2022automatic} and the broad applicability of pre-trained models. Nonetheless, a few studies focus on natural images captured by smartphone cameras for ubiquitous screening. \cite{darvariu2020quantifying} developed an application where users record their emotional state through photographs taken by the rear camera of smartphones. Alternatively, front-facing cameras can facilitate the capture of facial images, whose analysis can offer valuable insights into a person's emotional state \citep{wang2015using}. In recent work, \cite{nepal2024moodcapture} introduced a depression screening approach named MoodCapture, which was evaluated on a large dataset of facial images collected in the wild (over 125,000 photos). A key aspect of MoodCapture is the passive collection of images in multiple shots while users complete a self-assessment depression questionnaire on their smartphones. The authors argue that these images are preferable to traditional selfies as they capture more authentic and unguarded facial expressions.

Similarly to MoodCapture, the proposed work focuses on analyzing selfies captured by smartphone cameras for anxiety and depression screening. However, our study differs from MoodCapture in two major aspects. First, our work targets a specific population: high-risk pregnant patients accompanied by a team of healthcare professionals. This imposes a limitation on the dataset size when compared to crowd-sourced data. Second, our methodology focuses on face-centric selfies rather than analyzing general image aspects. To the best of our knowledge, this is the first study to explore the use of face-centric selfies for anxiety and depression screening in high-risk pregnant patients. Another relevant aspect of our work is the use of vision-language models (VLMs). Unlike traditional models that predict a limited set of basic emotions, VLMs can capture more nuanced emotions such as awe, shame, and emotional suppression \citep{bian2024understanding}. Despite being used for general emotion analysis, the particular application of VLMs for mental health screening is another contribution of our work.

\begin{figure}[t]
    \centering
    \includegraphics[width=0.75\columnwidth]{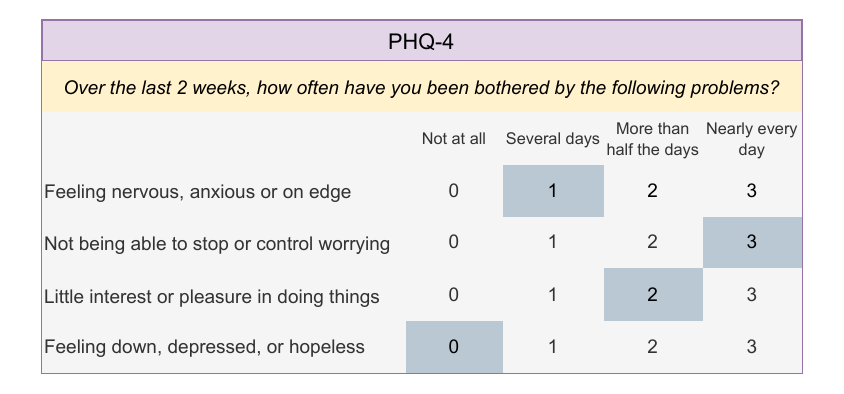}
    \caption{Patient Health Questionnaire-4 (PHQ-4) items and the respective frequency scores.}
    \label{fig:phq-4}
\end{figure}

\section{AI-Driven Screening via Selfie Analysis}
\label{sec:method}

This study relies on selfies taken by pregnant patients paired with their responses to the Patient Health Questionnaire-4 (PHQ-4), a four-item instrument for brief screening of anxiety and depression. The PHQ-4 was chosen for this study due to its efficiency and brevity in screening for both anxiety and depression, making it particularly useful in clinical settings where time is limited. Comprising only four items (Figure \ref{fig:phq-4}), it integrates questions from the PHQ-2 and GAD-2, which are validated tools for detecting depression and anxiety, respectively. This dual focus allows for a rapid assessment of two of the most prevalent mental health conditions, while maintaining strong psychometric properties, making it a reliable and practical tool for screening purposes in a clinical sample. The total score, calculated as the sum of individual scores, ranges from 0 to 12, with higher scores indicating greater severity of anxiety and depression symptoms. A score of 6 or higher on the PHQ-4 is typically used as a cut-off point for identifying cases where either anxiety or depression (or both) may be present and warrant further clinical evaluation \citep{caro2024systematic}. This cut-off point helps to identify individuals with moderate to severe symptoms of either condition, ensuring efficient screening, including mixed anxiety-depression states, which are fairly common among pregnant women. It allows for initial assessment without the need to differentiate between the two disorders \citep{javadekar2023biopsychosocial}.

\begin{figure}[t]
    \centering
    \includegraphics[width=\textwidth]{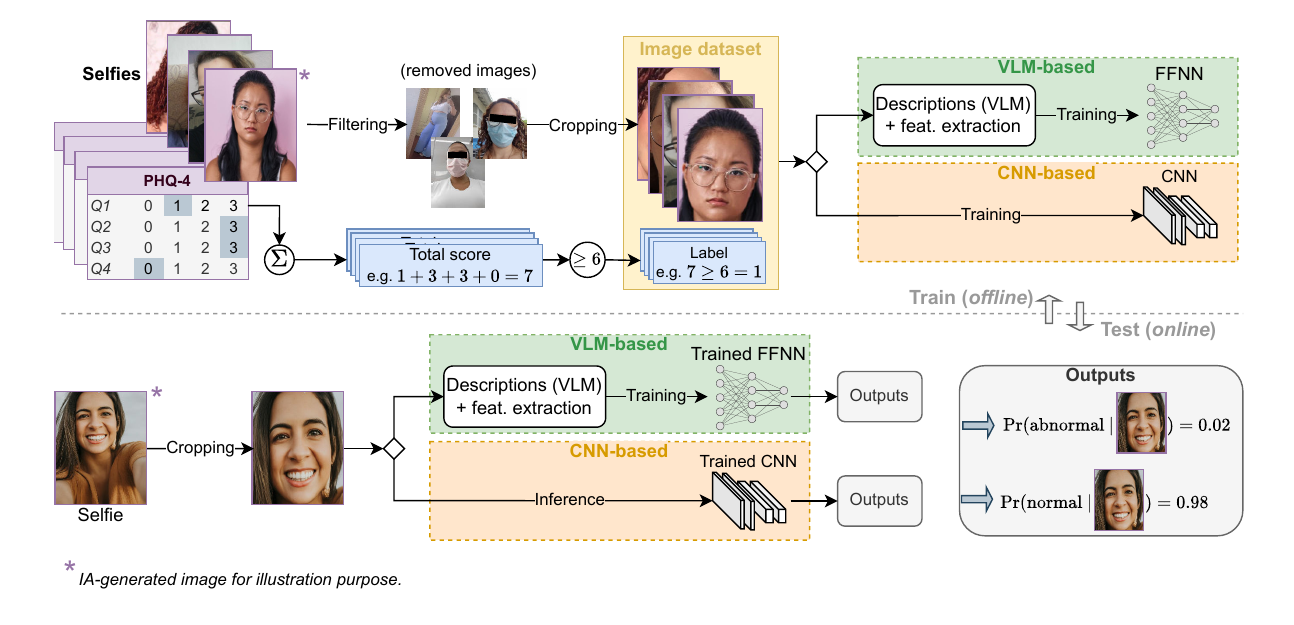}
    \caption{Overview of the proposed methodology for depression-anxiety detection using selfies and PHQ-4 responses. The training pipeline (top flow) involves filtering invalid selfies, cropping face regions with MTCNN, and labeling images based on PHQ-4 scores. The model (CNN or FFNN) is trained either directly from images (CNN-based) or text descriptions (VLM-based). The test pipeline (bottom flow) uses the trained model to classify new selfies.}
    \label{fig:overview}
\end{figure}

To enable AI-driven screening, we propose a methodology that leverages selfies and PHQ-4 responses to train machine learning (ML) models for depression-anxiety detection. Figure \ref{fig:overview} provides a joint overview of the VLM- and CNN-based approaches addressed in this work. There are two pipelines: the training pipeline (top flow) and the test pipeline (bottom flow). The training pipeline leverages image and PHQ-4 data to train an ML model for depression-anxiety detection. The training requires an image dataset comprising faces and the respective labels (0-normal, 1-abnormal). To assemble this dataset, the first step is manually filtering out invalid selfies, which are those taken with the rear-facing camera or with faces covered by a mask. The face region of the remaining samples is subsequently cropped by using a multi-task cascaded convolutional network (MTCNN) \citep{Zhang2016}. Selfies with multiple detected faces are discarded, as the PHQ-4 responses are individual. As previously explained, a label is derived for each face image by thresholding the PHQ-4 overall score.

With a valid set of face images and labels, an ML model can be trained. In the CNN-based approach, the face images are directly input into the classification model. The VLM-based approach, by contrast, involves generating descriptions of emotional states through face analysis and then extracting features from these textual descriptions. These features are subsequently used as inputs for the classification head, specifically a feed-forward neural network (FFNN). Once the model (CNN or FFNN) is trained, it can be used to classify an input selfie, as illustrated in the bottom flow of Figure \ref{fig:overview}. The face region in the selfie is cropped using the MTCNN, as in the top flow. The face image (or text features) is then input to the trained model, which outputs class probabilities. The depression-anxiety condition is verified if, and only if, $\operatorname{Pr}(\text{abnormal}\,|\,\text{sample}) > 0.5$. More details of the two detection approaches are provided in the following sections.

\subsection{CNN-based Approach}

This approach employs CNNs for predicting depression-anxiety directly from image data, as illustrated in Figure \ref{fig:overview}. Training from scratch is unfeasible in our context due to the reduced number of selfies/PHQ-4 responses: 147 samples from a total of 108 participants. To circumvent this issue, we start with CNN models pre-trained on the ImageNet dataset and fine-tune them on large-scale facial expression recognition (FER) datasets, which are closely related to our target task. A second fine-tuning is performed to adapt the FER pre-trained models to our task.

This work investigates the adaptation of models pre-trained on the FER2013 \citep{goodfellow2013challenges}, and RAF-DB \citep{li2017reliable} datasets, both of which encompass seven basic emotions (classes): anger, disgust, fear, happiness, sadness, surprise, and neutral state. Four CNN architectures were investigated: EfficientNetV2 \citep{Effnet2021}, ResNet-18 and ResNet-50 \citep{he2016deep}, and VGG11 \citep{VGG2015}. In the second fine-tuning stage, the classifier consists of a CNN backbone pre-trained on FER or RAF-DB appended with a 2-output fully connected layer (classification head). The backbone is frozen during training to prevent overfitting, ensuring that only the classification head is trained.

\begin{figure}[t]
    \centering
    \includegraphics[width=0.8\textwidth]{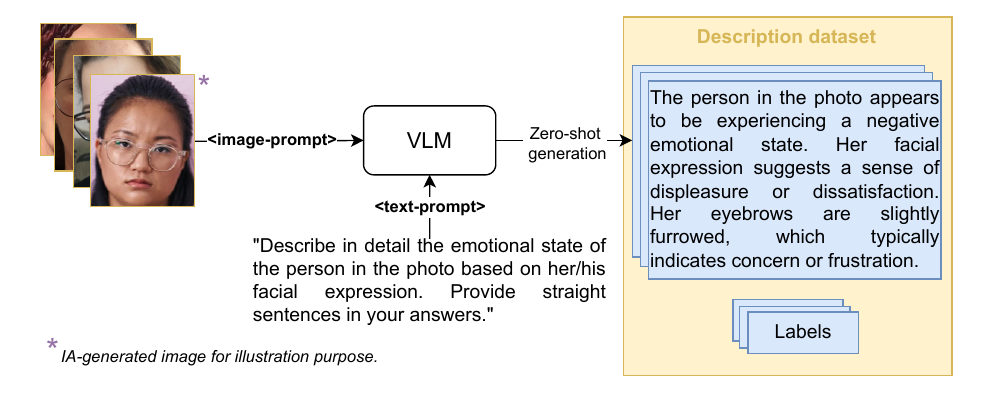}
    \caption{Zero-shot description generation with VLMs. The VLM prompt consists of an image (cropped face from a selfie) and a text instruction (text prompt). A description is generated for each face image in the image dataset. The label from each source image is transferred to the respective generated description, giving rise to an annotated dataset of textual descriptions.}
    \label{fig:gen-descriptions}
\end{figure}

\subsection{VLM-based Approach}

This approach leverages large generative vision-language models (VLMs) for facial analysis. Modern VLMs \citep{bordes2024introduction} combine visual encoders with large language models (LLMs) and can simultaneously learn from images and text. This enables them to perform various tasks, such as answering visual questions and captioning images. In particular, we benefit from the VLM zero-shot instruction-following ability to produce high-quality descriptions when provided with an image and text prompt. Three modern VLMs were investigated in this work: LLaVA-NeXT \citep{liu2024llavanext}, which is an improvement on LLaVA \citep{liu2024visual}, Kosmos-2 \citep{peng2023kosmos}, and the proprietary GPT-4o.

Figure \ref{fig:gen-descriptions} illustrates the intended usage of VLMs in this work. Instead of using the image-labeled dataset directly, textual descriptions (in the form of sentences) are generated by following the instruction in the input prompt: \emph{``Describe in detail the emotional state of the person in the photo based on her/his facial expression. Provide straight sentences in your answer.''}. It is worth mentioning that the prompt does not address the target task directly, i.e., detecting anxiety and/or depression. Nonetheless, experimental results (Section \ref{sec:results}) reveal that the pre-trained VLMs were unable to directly predict the depression-anxiety condition. This motivated using VLMs as analysts rather than judges, delegating the final decision to a secondary model, specifically an FFNN.

To build the classification model, features are extracted from the text descriptions by using a pre-trained Sentence-BERT \citep{reimers2019sentence} model called \texttt{all-MiniLM-L6-v2}, which is designed to embed sentences and paragraphs into a 384-dimensional representation. Two FFNN architectures (default and alternative) are investigated in this work, as expressed in the Equation \eqref{eq:nn}:

\begin{equation}
\label{eq:nn}
f_{net}(\textbf{x}) =
\begin{cases}
 \bm{W}\bm{x} + \bm{b} &\text{(default)} \\
 \bm{W}_2(\operatorname{ReLU}(\bm{W}_1\bm{x} + \bm{b}_1) + \bm{b}_2 &\text{(alternative),}
\end{cases}
\end{equation}
where $\bm{W}_{\bigcdot}$ and $\bm{b}_{\bigcdot}$ denote a weight matrix and a bias vector, respectively. The default model is a simple linear classifier, while the alternative model includes a hidden layer and is explored in a sensitivity analysis in the experiments. The logits of the network, $f_{net}(\textbf{x})$, are used to calculate the class probabilities through softmax normalization: $\operatorname{Pr}(\bigcdot\,|\,\textbf{x}) = \operatorname{softmax}(f_{net}(\textbf{x}))$.

\begin{figure}[t]
    \centering
    \includegraphics[width=0.75\columnwidth]{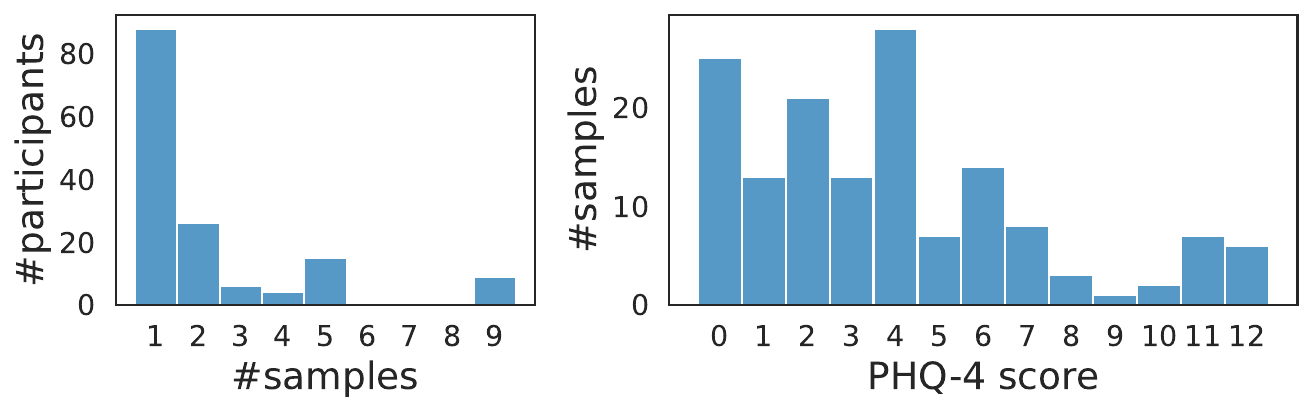}
    \caption{Data distribution after removing invalid samples. Most subjects contributed with a single sample, while one contributed with nine. The imbalance is evident, with a bias towards negative samples (PHQ-4 \,$<$\, 6).}
    \label{fig:data-distribution}
\end{figure}

\section{Experimental Methodology}
\label{sec:experimental-methodology}

This section outlines the elements of the experimental methodology: the data collection procedure involving selfies and PHQ-4 responses, the experiments conducted, and, finally, the hardware and software setup.

\subsection{Data Collection}
\label{sec:data}

Selfies and PHQ-4 responses constitute a subset of data collected for a research project (local
ethics board consent: CAAE 64158717.9.0000.0065) involving high-risk pregnant patients from the Clinical Hospital of the University of São Paulo (São Paulo, Brazil). The research focused on pregnant women aged 18 and above who could provide informed consent, had at least an elementary education, and owned a smartphone. Data collection utilized smartphones, with the participants capturing natural photos (selfies) in such a way that their faces were visible. They were responsible for uploading image data to the REDCap platform \citep{harris2009research} instance hosted at the University of São Paulo, which was also utilized for completing the PHQ-4 questionnaire.

For this study, 108 participants were recruited and instructed to submit bi-weekly reports, each consisting of the PHQ-4 questionnaire\footnote{The Portuguese version of the PHQ-4 used in this work is provided by Pfizer (Copyright\textcopyright, 2005 Pfizer Inc).} and a selfie. Figure \ref{fig:data-distribution} illustrates the data distribution after manually removing invalid selfies. The left chart displays a total of 147 selfies from the 108 participants. Although each participant was expected to contribute multiple samples (11 in total), most participants submitted only a single selfie. The right-side chart highlights the imbalance in the data, with a bias towards negative samples (PHQ-4 \,$<$\, 6): 106 negative versus 41 positive samples.

\subsection{Comparative Evaluation}

The comparative evaluation aims to assess the performance of the CNN- and VLM-based approaches for detecting depression-anxiety in selfies. Recall that while a CNN architecture is trained in the first approach, a FFNN architecture is trained in the latter approach. The performance of pre-trained VLMs for zero-shot screening was also evaluated to serve as a baseline for the VLM-based approach. To conduct the experiments, the collected data was processed to create an image dataset consisting of cropped faces and labels, as discussed in Section \ref{sec:method}. The CNN and FFNN models were trained and tested under a \emph{Leave One Subject Out} (LOSO) cross-validation protocol to avoid performance overestimation. This protocol ensures that each subject's data is used as a unique test set while the remaining data forms the training set, providing a robust measure of model performance across different individuals.

In each training-testing session, the samples of a single subject are classified, and the predictions (normal or abnormal) are recorded. Once all predictions are available, performance metrics are calculated. Traditional metrics in binary classification are utilized: precision, recall, F1-score, area under ROC curve (AUC), and accuracy. F1-score is particularly relevant because it accounts for the dataset imbalance, a characteristic of our data. Specific implementation details for each depression-anxiety detection approach are provided below. 

\paragraph{CNN-based Approach} The pre-training on both FER2013 and RAF-DB was conducted for 30 epochs with a fixed learning rate of $10^{-4}$ and a $\ell$2-regularization factor of $10^{-5}$. In the default settings, fine-tuning was conducted for 30 epochs, with a learning rate of $10^{-4}$, and an $\ell$2-regularization factor of $10^{-5}$. Different parameters were used based on empirical evidence of overfitting in the training. ResNet-18 (RAF-DB): for 50 epochs, a learning rate of $2 \times 10^{-5}$, and a weight decay of $10^{-6}$. ResNet-50 (RAF-DB): it included a dropout regularization with a probability of 50\%. Pre-training and fine-tuning utilized Adam optimization. 

\paragraph{VLM-based Approach} VLMs were configured for deterministic inference, employing a strategy that selects the most probable next token at each step of the generation process. The default (linear) model (Equation \ref{eq:nn}) was adopted as the classifier. To address the dataset imbalance, the descriptions associated with positive samples (minority class) were upsampled to match the number of negative samples. For each VLM, the classification head underwent training using the Adam optimizer for 15 epochs. Additional training parameters included a learning rate of $10^{-4}$, a batch size of 2, and a $\ell$2-regularization factor of $10^{-4}$. During each training session, a randomly selected subset comprising 10\% of the training data was reserved for validation purposes. The best-epoch checkpoint was determined based on achieving the highest F1-score on the validation set.

\paragraph{Zero-shot Screening with VLMs} A natural question arises when using powerful models such as VLMs: ``Are pre-trained VLMs capable of zero-shot screening depression-anxiety?'' To answer this question, we conducted an experiment in which VLMs were asked to classify an input image based on the following instruction prompt: \emph{``Describe in detail the emotional state of the person in the photo based on her/his facial expression. Provide straight sentences in your answers. Based on your description, classify the emotional state as either 'normal', 'anxiety', or 'depression'. The output must be exactly one of these words. Follow the template: Output: \{result\}''.}
In preliminary tests, only GPT-4o followed strictly the instruction prompt, while LLaVA-NeXT and Kosmos-2 generated longer and more complex descriptions. Therefore, this experiment was performed only for GPT-4o. GPT-4o classified each of the 147 samples into one of the three classes: normal, anxiety, or depression. The samples classified as anxiety or depression were considered positive, while the normal samples were considered negative.

\subsection{Sensitivity Analysis}

This additional experiment explores the effects of incorporating a hidden layer into the FFNN of the VLM-based approach (alternative model, Equation \ref{eq:nn}). The model was evaluated with various hidden units: $h=4, 8, 16, \ldots, 256$. This investigation was motivated by the promising results observed in the VLM-based approach, as elaborated further in Section \ref{sec:results}. Furthermore, understanding the impact of the classification head is relevant because it is the only trainable component in the VLM-based pipeline. The LOSO protocol was also employed in this investigation. Additionally, during model training, a dropout layer was incorporated after the ReLU activation function to prevent overfitting.

\subsection{Hardware-software Setup}

Hardware: Intel(R) Xeon(R) CPU @ 2.20GHz with 32GB of RAM, running Linux Ubuntu 22.04.4 LTS, and equipped with an NVIDIA L4 GPU with 24GB of memory. Software: The source code was written in Python 3.10, mostly using PyTorch 2.3 for model training and inference. The Sentence-BERT \citep{reimers2019sentence} is implemented in SentenceTransformers library\footnote{\url{https://sbert.net}}. The LLaVA-NeXT and Kosmos-2 models -- \texttt{llava-hf/llava-v1.6-mistral-7b-hf} and \texttt{microsoft/kosmos\-2-patch14-224}, respectively, are available at the HuggingFace\footnote{\url{https://huggingface.co}}, while GPT-4o was accessed via the OpenAI API.
\section{Results and Discussion}
\label{sec:results}

This section presents the results of the conducted experiments: comparative evaluation and sensitivity analysis. Limitations and challenges are also discussed at the end of this section.

\begin{table}[t]
\centering
\caption{Overall results for the comparative evaluation (\%).}
\label{tab:main}
\resizebox{0.8\textwidth}{!}{%
\begin{tabular}{l|c|rrrrr}
\toprule
\textbf{Model} & \textbf{Pre-train} & \textbf{Prec.} & \textbf{Rec.} & \textbf{F1-score} & \textbf{AUC} & \textbf{Acc.} \\
\midrule
\multicolumn{7}{l}{\textbf{CNN-based}}\\
\midrule
ResNet-18 & \multirow{4}{*}{FER2013} & 36.2 & 51.2 & 42.4 & 62.7 & 61.2 \\
ResNet-50 & & 36.5 & 46.3 & 40.9 & 60.1 & 62.6 \\
VGG11 & & 39.0 & 39.0 & 39.0 & 57.1 & 66.0 \\
EfficientNetV2 & & 39.6 & 51.2 & 44.7 & 65.7 & 64.6 \\ 
\midrule
ResNet-18 & \multirow{4}{*}{RAF-DB} & 37.9 & \textbf{53.7} & 44.4 & 64.5 & 62.6 \\
ResNet-50 & & 28.0 & 34.1 & 30.8 & 52.1 & 57.1 \\
VGG11 & & 28.6 & 43.9 & 34.6 & 54.8 & 53.7 \\
EfficientNetV2 & & \textbf{41.7} & 48.8 & \textbf{45.0} & \textbf{68.4} & \textbf{66.7} \\
\midrule
\multicolumn{7}{l}{\textbf{VLM-based}}\\
\midrule
GPT-4o$^\star$ & \multirow{3}{*}{-} & \textbf{61.8} & 51.2 & \textbf{56.0} & \textbf{72.6} & \textbf{77.6} \\
Kosmos-2$^\star$ & & 40.0 & \textbf{53.7} & 45.8 & 72.0 & 64.6 \\
LLaVA-NExT$^\star$ & & 41.2 & 51.2 & 45.7 & 66.2 & 66.0 \\
\midrule
\multicolumn{7}{l}{\textbf{Zero-shot Screening}}\\
\midrule
GPT-4o & - & 53.9 & 34.2 & 41.8 & 61.4 & 73.5 \\
\bottomrule
\multicolumn{7}{l}{{\raggedright \textbf{Bold} values indicate the highest metric value within each approach.}} \\
\multicolumn{7}{l}{{\raggedright $^\star$VLM with (default) linear classifier model.}} \\
\end{tabular}
}
\end{table}

\subsection{Comparative Evaluation}

Table \ref{tab:main} shows the results of the comparative evaluation experiment. Overall, EfficientNetV2 outperformed the compared models among the CNNs for both FER2013 and RAF-DB pre-training, while GPT-4o yielded the best performance among the VLMs. Pre-training on RAF-DB improved most metrics for EfficientNetV2 and ResNet-18, although a notable decrease in performance was observed for ResNet-50 and VGG11. Notably, ResNet-18 yielded better performance than ResNet-50, despite its reduced size. This suggests that larger models might require more data to achieve better performance.

Regarding the F1-score, using RAF-DB resulted in only a 0.3 p.p. improvement over FER2013 for EfficientNetV2, while recall decreased by 2.4 p.p.. In medical applications, lower recall indicates the risk of missing a significant number of actual cases, which can lead to undiagnosed conditions and potentially severe consequences for patient health and safety. In this context, a remarkable gain was observed for ResNet-18, whose recall raised from 46.3 to 53.7\% with RAF-DB. Despite its lower precision compared to EfficientNetV2, ResNet-18 presents itself as a viable alternative given the critical importance of the recall metric in this context. Moreover, its F1-score is only marginally lower by less than 1 p.p. compared to EfficientNetV2. The use of CNNs represents a more traditional way to address this problem. The obtained results for these models reveal how challenging the task is for the addressed scenario, specifically when using a small dataset.

As an alternative, we proposed using pre-trained VLMs due to their ability to analyze faces. Table \ref{tab:main} shows the results for the VLM-based approach with the classification head (FFNN) in its default configuration. Overall, the F1-score for the three evaluated VLMs surpassed the CNN-based models, with GPT-4o achieving the highest value. GPT-4o outperformed the open-source models, Kosmos-2 and LLavA-NExt, across all metrics, except for the recall metric, where Kosmos-2 achieved the same 53.7\% as ResNet-18. LLAvA-NExT and Kosmos-2 showed a similar F1-score, with the most significant difference observed in the AUC metric: nearly 6 p.p. in favor of Kosmos-2. A particularly notable result is the F1-score obtained with GPT-4o, which surpasses its competitors by a large margin, almost 10 p.p. higher, being the only model to perform above 50\% in this metric. In summary, GPT-4o yielded the best performance considering both CNN- and VLM-based approaches.

\begin{figure}
    \centering
    \includegraphics[width=0.7\textwidth]{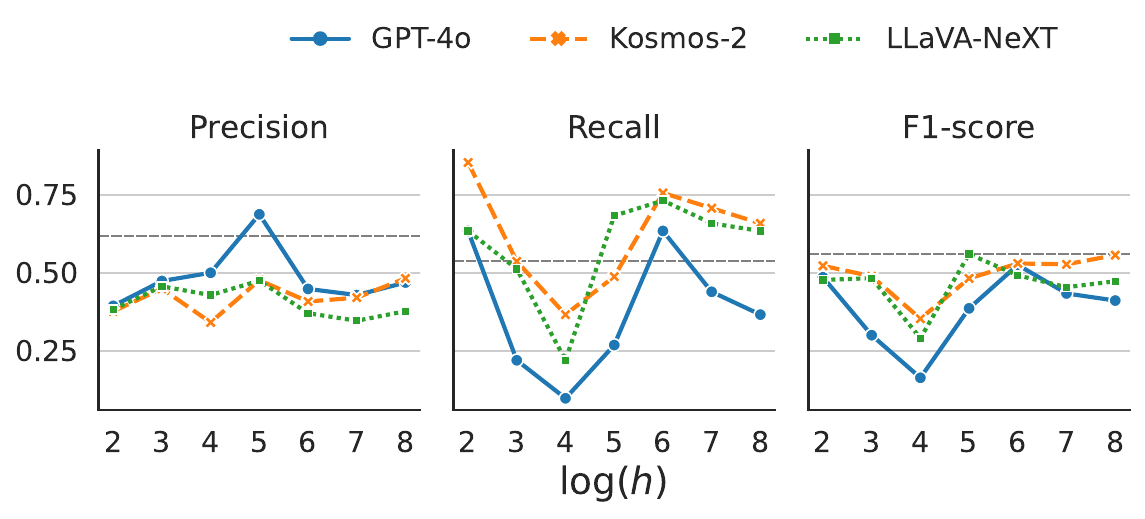}
    \caption{Sensitivity analysis. The FFNN classifier (VLM-based) was evaluated with various hidden units: $h=4, 8, 16, \ldots, 256$. The dashed line in each chart represents the highest value for the respective metric as reported in Table \ref{tab:main}.}
    \label{fig:sensitivity}
\end{figure}

The last row of Table \ref{tab:main} shows the results for the zero-shot screening with GPT-4o. While the accuracy in this scenario was superior to that obtained with LLAvA-NExT and Kosmos-2, the recall and F1-score demonstrated the opposite trend. Remarkably, the recall in zero-shot screening was 17 p.p. lower than that obtained with the VLM-based classification models. By comparing with the GPT-4o in the VLM-based approach, we conclude that the proposed formulation based on description generation and classification is crucial for achieving high performance.

\subsection{Sensitivity Analysis}

While Table \ref{tab:main} focused on the default FFNN model (VLM-based), this experiment investigated the alternative (single hidden layer) FFNN model (Equation \eqref{eq:nn}). More specifically, it was analyzed 
the impact of increasing the model complexity by varying the number of hidden units ($h=4, 8, 16, \ldots, 256$). Figure \ref{fig:sensitivity} shows the results of this experiment. The dashed line in each chart represents the highest value for the respective metric as reported in Table \ref{tab:main}. Notably, the F1-score achieved with GPT-4o and the default classification model (56.0\%) -- indicated by the dashed line in the third chart (from left to right) -- demarcates an empirical upper bound for this metric.
 
The F1-score curves follow a similar pattern to the recall curves, both exhibiting a `V' shape from $h=4$ to $64$. Kosmos-2 outperformed the other VLMs in recall in most cases, achieving a maximum recall of 85.3\% for $h=4$ (with an F1-score of 52.2\%). The high recall indicates a solid ability to flag more potential cases, which is highly desirable for this type of application. Kosmos-2 (with $h=256$) and LLAvA-NExT (with $h=32$) were able to match the F1-score of GPT-4o with the default model, however, with significantly higher recall: 65.3 and 68\% for Kosmos-2 and LLAvA-NExT, respectively, compared to 51.2\% achieved by GPT-4o. This shows that open-source VLMs can yield competitive performance when combined with a more complex classification model.

\begin{figure}[t]
    \centering
    \subfloat[PHQ-4\,=\,8]{\label{fig:limitation-a}\includegraphics[width=0.2\textwidth,height=3cm]{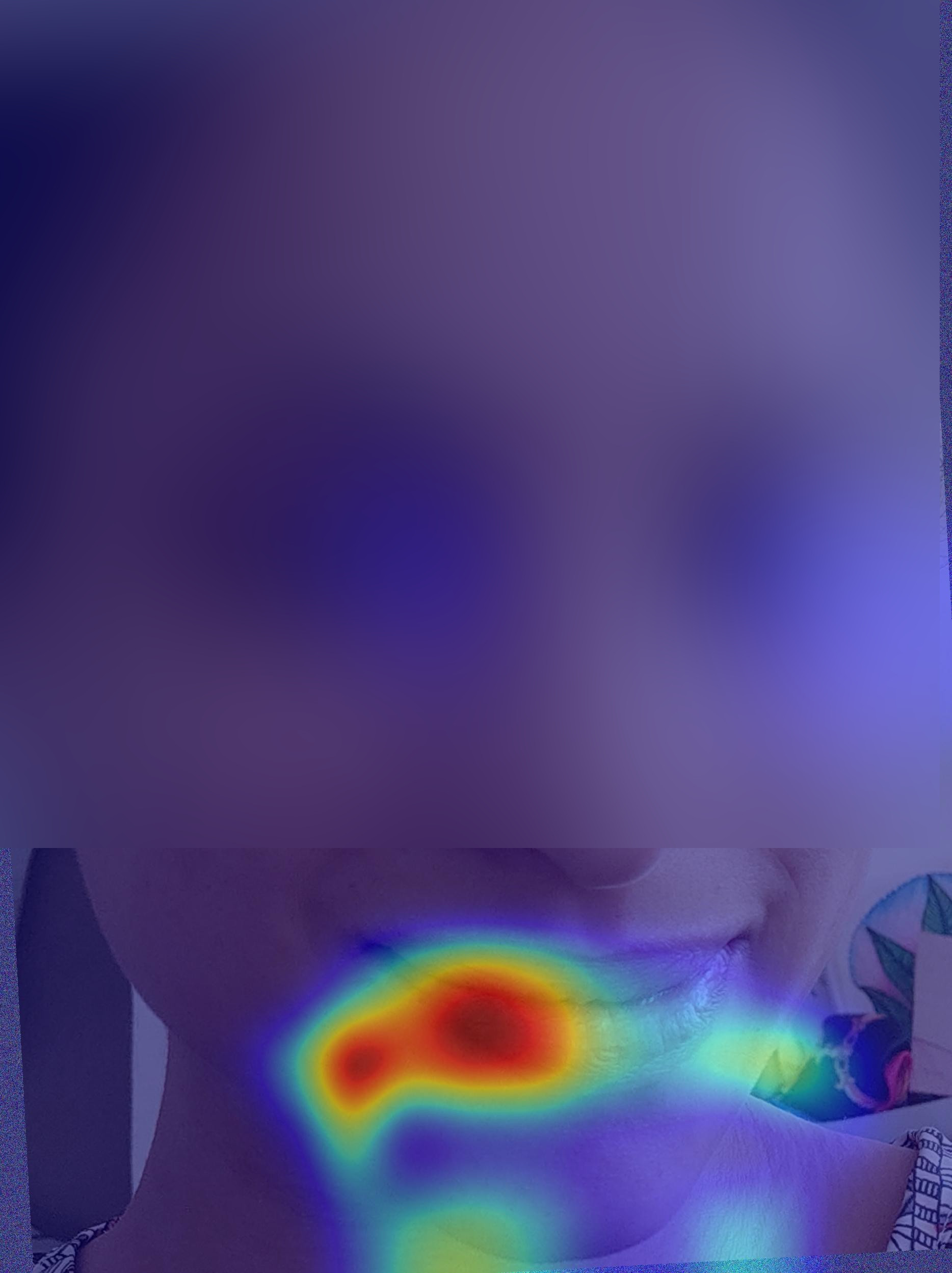}}
    \subfloat[PHQ-4\,=\,2]{\label{fig:limitation-b}\includegraphics[width=0.2\textwidth,height=3cm]{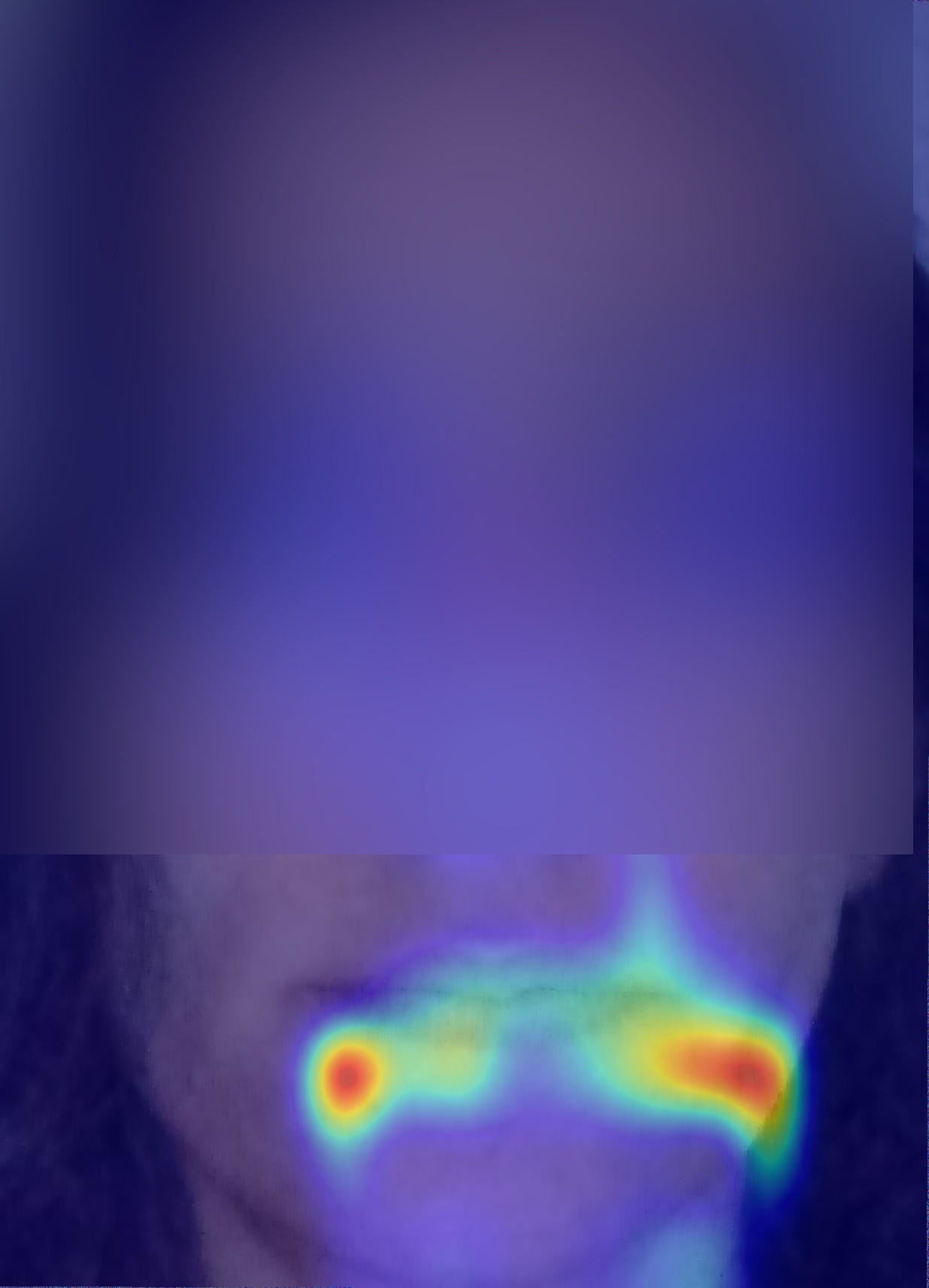}}
    \subfloat[PHQ-4\,=\,7]{\label{fig:limitation-c}\includegraphics[width=0.2\textwidth,height=3cm]{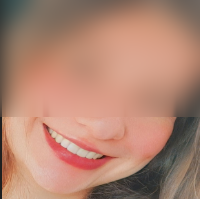}} 
    \subfloat[PHQ-4\,=\,2]{\label{fig:limitation-d}\includegraphics[width=0.2\textwidth,height=3cm]{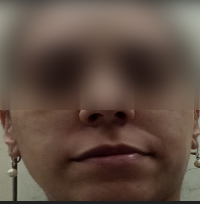}}
    \caption{Challenging cases that resulted in misclassification. The Grad-CAM attention maps highlight the areas around the mouth as the most influential for EfficientNetV2's prediction in (a) and (b).}
    \label{fig:limitation}
\end{figure}

\subsection{Limitations and Challenges}

Learning the relationship between facial expressions and self-reported PHQ-4 scores from single selfies can be challenging with small datasets. When using CNNs, prediction is mainly influenced by features around the mouth, as evidenced by the Grad-CAM \citep{selvaraju2017grad} attention maps for EfficientNetV2 in Figures \ref{fig:limitation-a} and \ref{fig:limitation-b}. In Figure \ref{fig:limitation-a}, the slight smile led the model to classify a positive sample (PHQ-4\,$\geq$\,6) as negative. However, facial cues around the eyes might suggest a posed (non-Duchenne) smile, which is not necessarily a sign of happiness or well-being. This issue could be mitigated by incorporating facial action units (AUs) into the learning process, enhancing attention guidance to other critical regions, as recently proposed in a study on basic emotions \citep{belharbi2024guided}.

Approximately 12\% of the errors with GPT-4o involve positive samples misclassified as negative where the individual is smiling. Figure \ref{fig:limitation-c} shows an example of a genuine (Duchenne) smile associated with a PHQ-4 score of 7 (close to the threshold) that was misclassified as negative. The description generated by GPT-4o includes terms such as ``The person in the photo appears to be happy'', ``She is smiling broadly'', and ``Her eyes are slightly squinted'', which accurately reflect the visible facial expressions. This situation is potentially misleading even for experienced face analysts, as the individual's mood is positive, but the PHQ-4 score is close to the threshold.

Nearly 18\% of the errors with GPT-4o are related to negative samples with PHQ-4 score lower than 3 described by the model as ``neutral''. The sample in Figure \ref{fig:limitation-d} (PHQ-4\,=\,2) was described by GPT-4o as ``neutral or somewhat weary expression'', ``corners of their mouth are slightly downturned'', ``lack of enthusiasm''. This suggests that the negative elements critically influenced the positive response despite the neutral emotional state described by the VLM. From a data perspective, multiple captures or fusion with complementary data modalities could yield a significant improvement in the overall performance for both smiling and neutral expressions scenarios.

AI systems are notorious for showing errors and biases in different racial groups \citep{nazer2023bias}. Despite the significance of this issue and its implications for fairness and equity, this sensitive topic was not explored in the present study.
\section{Conclusion}
\label{sec:conclusion}

This work addressed the AI-driven mental health screening in mobile applications using face-centric selfies. The scope of the study included collecting a dataset of selfies paired with responses to a self-reported questionnaire (PHQ-4) and evaluating two depression-anxiety detection approaches. In the comparative evaluation, the proposed VLM-based approach yielded better results than the typical transfer learning with CNNs or zero-shot screening with GPT-4o. Notably, GPT-4o achieved the best F1-score (56\%) and accuracy (77.6\%), while Kosmos-2 attained the highest recall (53.7\%).

The sensitivity analysis demonstrated that open-source VLMs can yield competitive performance, nearing the F1-score of GPT-4o. When combined with more complex FFNNs, Kosmos-2 and LLAvA-NExT achieved 56\% of F1-scores but with significantly higher recall, which is particularly important in this context. Specifically, Kosmos-2's recall increased to 85.3\% (with an F1-score of 52.2\%) by adding a 4-unit hidden layer.

Future work will explore two main directions. From a data perspective, we plan to collect a larger dataset with more reliable data by developing a smartphone application for multiple-shot passive capture of face images. From a methodological perspective, smile analysis and action unit information will be investigated to address the limitations of the current approach. Furthermore, fusion with complementary data modalities, such as audio and text transcriptions, will be investigated to enhance the screening performance.


\bibliographystyle{abbrvnat}
\bibliography{paper} 

\end{document}